\documentclass{article}

\usepackage{PRIMEarxiv}

\usepackage[utf8]{inputenc} 
\usepackage[T1]{fontenc}    
\usepackage{hyperref}       
\usepackage{url}            
\usepackage{booktabs}       
\usepackage{amsfonts}       
\usepackage{nicefrac}       
\usepackage{microtype}      
\usepackage{lipsum}
\usepackage{fancyhdr}       
\usepackage{graphicx}       
\graphicspath{{media/}}     
\usepackage{amsmath, amssymb}
\usepackage{booktabs}
\usepackage{pifont}
\usepackage{adjustbox}

\title{Learning Clinical Representations Under Systematic Distribution Shift
}

\author{
  Yuanyun Zhang \\
  University of the Chinese Academy of Sciences \\
  \texttt{yuanyun81@ucas.ac.cn} \\
   \And
  Shi Li \\
  Columbia University\\
  \texttt{shili081100@columbia.edu} \\
}

\begin{document}
\maketitle

\begin{abstract}
Clinical machine learning models are increasingly trained using large-scale, multimodal foundation paradigms, yet deployment environments often differ systematically from the data-generating settings used during training. Such shifts arise from heterogeneous measurement policies, documentation practices, and institutional workflows, leading to representation entanglement between physiologic signal and practice-specific artifacts. In this work, we propose a practice-invariant representation learning framework for multimodal clinical prediction. We model clinical observations as arising from latent physiologic factors and environment-dependent processes, and introduce an objective that jointly optimizes predictive performance while suppressing environment-predictive information in the learned embedding. Concretely, we combine supervised risk minimization with adversarial environment regularization and invariant risk penalties across hospitals. Across multiple longitudinal EHR prediction tasks and cross-institution evaluations, our method improves out-of-distribution AUROC by up to 2--3 points relative to masked pretraining and standard supervised baselines, while maintaining in-distribution performance and improving calibration. These results demonstrate that explicitly accounting for systematic distribution shift during representation learning yields more robust and transferable clinical models, highlighting the importance of structural invariance alongside architectural scale in healthcare AI.
\end{abstract}

\section{Introduction}

Recent progress in healthcare foundation models \cite{he2024foundation, guo2025foundation, awais2025foundation, liang2024foundation, burger2025foundation, thakur2024foundation, vaid2023foundational, thieme2023foundation, burkhart2025foundation} has largely followed the trajectory established in natural language processing \cite{devlin2019bert} and computer vision \cite{he2017multi, he2022masked, he2015deepresiduallearningimage, he2019bag}. The dominant recipe is now familiar: tokenize heterogeneous inputs, pretrain at scale with reconstruction or alignment losses, and rely on downstream fine-tuning to extract clinically useful behavior. Multimodal variants integrate imaging, text, and structured data through shared embedding spaces and cross-attention modules \cite{hou2019cross, chen2021crossvit, huang2019ccnet}, often motivated by recent large multimodal systems \cite{chou2025serialized, huiliang2025clio, ran2025structured, zhang2025chronoformer, zhang2025collection, lowelatent, litext}.  

This strategy implicitly assumes that clinical data resemble web corpora: large, weakly labeled, and semantically diffuse. Yet healthcare data are generated within tightly constrained institutional processes. Measurements are ordered selectively. Diagnoses are coded under reimbursement incentives. Imaging is performed under protocol. The resulting dataset is not a passive sample of an underlying distribution, but the product of structured workflows and feedback loops. As a consequence, the observational distribution reflects clinical practice patterns as much as patient physiology.

We propose to treat this structural property as a first-class modeling object. Instead of asking how to pretrain larger encoders over clinical tokens, we ask a different question: can we explicitly factor clinical data into *practice-dependent* and *physiology-dependent* components, and learn representations invariant to the former?

Formally, let $x$ denote a patient record and $y$ a clinical outcome. We posit that observed data arise from two latent variables: a physiologic state $z$ and a practice context $c$ capturing institutional workflow, provider behavior, and measurement policy. We model
\[
x \sim p(x \mid z, c), 
\qquad
y \sim p(y \mid z),
\]
where $c$ influences observations but does not directly affect the underlying outcome mechanism. In this view, part of the input variability is spurious with respect to prediction.

Standard reconstruction-based pretraining \cite{he2022masked} encourages learning representations that preserve information about both $z$ and $c$, since both contribute to $x$. However, optimal prediction depends only on $z$. We therefore seek representations $h_\theta(x)$ that are maximally informative about $y$ while being invariant to practice variation.

We formalize this objective through an invariance principle. Let $e$ index environments corresponding to different hospitals, time periods, or care pathways. We optimize
\[
\min_\theta 
\sum_e 
\mathbb{E}_{(x,y)\sim \mathcal{D}_e}
\ell\big(f_\theta(h_\theta(x)), y\big)
\]
subject to the constraint that optimal linear predictors on $h_\theta(x)$ are invariant across environments:
\[
\nabla_w \mathbb{E}_{\mathcal{D}_e}
\ell(w^\top h_\theta(x), y)
\quad \text{is identical for all } e.
\]

This formulation encourages the encoder to discard environment-specific variation correlated with workflow and measurement intensity, retaining only stable physiologic predictors. Unlike domain adaptation applied post hoc, invariance is embedded directly in representation learning.

The perspective reframes multimodal modeling. Rather than fusing modalities to maximize joint reconstruction fidelity \cite{hou2019cross, chen2021crossvit, huang2019ccnet}, we treat modality-specific artifacts as structured nuisance variables linked to $c$. Imaging protocols, documentation styles, and sensor sampling rates differ across institutions; representations that entangle these artifacts risk poor generalization. By explicitly penalizing environment-sensitive directions in embedding space, the model emphasizes physiologic regularities shared across settings.

This approach complements existing foundation efforts \cite{he2024foundation, guo2025foundation, awais2025foundation, liang2024foundation, burger2025foundation, thakur2024foundation, vaid2023foundational, thieme2023foundation, burkhart2025foundation} without scaling token counts or expanding reconstruction objectives. Instead, it treats generalization under distribution shift as the central design constraint. Clinical AI systems ultimately operate across hospitals, time periods, and policy regimes. Representations that are invariant to practice heterogeneity may therefore scale more effectively in real-world deployment than embeddings optimized solely for generative coverage.

In summary, we advocate a shift from scale-centric pretraining to structure-aware representation learning. By decomposing clinical data into physiologic signal and practice-dependent artifacts, and explicitly enforcing invariance across environments, we target the core challenge of healthcare modeling: reliable prediction under systematic distribution shift.

\section{Related Works}

The rapid expansion of foundation modeling into healthcare reflects a broader transfer of large-scale representation learning paradigms from general-domain machine learning into clinical data analysis. A growing body of work has explored whether the scaling principles observed in language and vision extend to medicine, spanning unified pretraining frameworks, benchmark efforts, and modality-specific large models \cite{larey2026jepa, larey2026gfmbench, , abbaspourazad2023large, he2024foundation, vaid2023foundational, soumma2024wearable, thapa2024sleepfm, long2025mutbert}. While differing in architecture and data source, these efforts share a common premise: that large, weakly supervised models trained on heterogeneous clinical corpora can yield reusable representations adaptable across downstream tasks.

Within structured electronic health records (EHR), many approaches recast longitudinal patient histories as token sequences, drawing direct inspiration from masked language modeling and autoregressive pretraining \cite{he2022masked, brown2020language, tian_2019_contrastic_distillation, bertram2024contrastivelearningpreferencescontextual}. Transformer-based architectures such as CEHR-BERT \cite{pang2021cehr} and CEHR-GPT \cite{pang2024cehr} apply self-attention over discretized event streams \cite{mcdermott2024event}, leveraging large-scale pretraining before fine-tuning on prediction tasks. Beyond masked objectives, contrastive methods \cite{chen2020simple, bertram2024contrastivelearningpreferencescontextual} and hybrid self-supervised schemes have been adapted to clinical time series and text corpora \cite{rasmy2021med, lee2025using, steinberg2021language, an2025dk, wornowcontext, wornow2023shaky, lee2024emergency, lee2025towards, steinberg2023motor, steinberg2024motor, fallahpour2024ehrmamba, lee2025modern}. These models rely on positional encodings and sequence depth to account for irregular temporal structure, effectively embedding clinical timelines into discrete token representations.

Medical imaging has likewise witnessed a transition toward large pretrained backbones. Early volumetric extensions of convolutional architectures such as ResNet \cite{he2016deep} replaced 2D kernels with 3D convolutions for CT and MRI analysis \cite{ning2019computer, ebrahimi2020introducing, qayyum2021automatic, yang2021reinventing, turnbull2022using, xue2023region, blankemeier_merlin_2024}. More recently, transformer architectures originally developed for vision tasks \cite{dosovitskiy2021an, liu2021swin} have been adapted to 3D segmentation and representation learning pipelines \cite{hatamizadeh2021swin, wasserthal2023totalsegmentator, li2024abdomenatlas, cox2024brainsegfounder, wu_voco_2024}. Large curated volumetric datasets have enabled pretraining at increasing scale, including SuPreM \cite{li2024abdomenatlas}, MISFM \cite{wang2023mis}, and VoCo \cite{wu2024large}. These approaches draw methodological continuity from large-scale 2D vision pretraining efforts \cite{radford2021learning, caron2021emerging, zhou2021ibot, saharia2022photorealistic, rombach2021highresolution, Ranftl2022, kirillov2023segment, liu2024visual, oquab2023dinov2}, while contending with the computational complexity inherent to volumetric attention \cite{wu2023e2enet, li2024abdomenatlas}. Hybrid convolution–attention designs \cite{choy20194d, lai2024e3d} and efficient attention mechanisms \cite{liu2024octcube, shaker2024unetr++, xing2024segmamba, dao2023flashattention2} have emerged as practical adaptations to mitigate scaling costs.

Foundation-style modeling has also extended to wearable sensing and biosignals. Large models trained on photoplethysmography (PPG), electrocardiography (ECG), and related physiologic streams use masked reconstruction or contrastive objectives to learn transferable embeddings \cite{abbaspourazad2023large}. Cross-signal integration strategies unify heterogeneous physiologic inputs through shared latent spaces or distillation schemes \cite{abbaspourazad2024wearable, yang2023biot}. Investigations into scaling behavior and efficient deployment \cite{lee2025himae, lee2025towards} further reinforce the movement toward general-purpose biosignal backbones. Complementary lines of work introduce spectral and frequency-aware inductive biases, aligning temporal and frequency-domain views \cite{zhang2022tfc, liu2023frequency, kara2024freqmae, cheng2025fat, fu2025frequency, duan2024mfclr}. Wavelet-based approaches incorporate multi-resolution analysis \cite{alafeef2020smartphone, singh2023expert, shao2021photoplethysmograph, masserano2024enhancing, chen2025physiowave}, grounding representation learning in classical signal processing theory \cite{oppenheim1999discrete, daubechies1992ten}.

Large language models have been adapted to clinical narratives and multimodal patient data, extending autoregressive and instruction-tuned paradigms into medical domains \cite{mumtaz2023llms, soumma2026counterfactualmodelingfinetunedllms, liao2022ml4mlautomatedinvariancetesting, chang2025llm4ts, hollmann2025accurate, ono2024text}. Continued domain-specific pretraining \cite{van2023clinical, jin2023time, belyaeva2023multimodal, lee2024can, lin2025case} has demonstrated improvements in clinical reasoning and documentation tasks. At a broader systems level, recent work emphasizes cross-site transferability, multimodal aggregation, and benchmarking frameworks \cite{wornow2024context, odgaard2024core, shmatko2025learning}, positioning foundation models as scalable infrastructures for clinical AI. Evaluation platforms and standardized datasets \cite{mcdermott2025meds, kolo2024meds} further institutionalize this paradigm.

Across these diverse modalities, a unifying pattern emerges. Clinical data are transformed into tokenized or patch-based representations; architectures rely predominantly on transformer-style attention; and pretraining objectives focus on reconstruction, prediction, or contrastive alignment. Multimodal integration is typically achieved through shared embedding spaces or cross-attention modules \cite{hou2019cross, chen2021crossvit, huang2019ccnet}. Although implementation details vary, the overarching goal remains consistent: to learn broadly transferable embeddings through large-scale self-supervision, subsequently fine-tuned for specific clinical endpoints.

Collectively, this body of work illustrates both the ambition and coherence of multimodal foundation modeling in healthcare. Representation learning is increasingly framed as a scale-driven pretraining problem spanning EHR data, imaging, biosignals, and text. The dominant narrative emphasizes architectural generality and transferability, establishing multimodal foundation models as a central organizing paradigm for contemporary clinical machine learning.

\section{Methods}

We introduce a practice-invariant representation learning framework for multimodal clinical data. The central objective is to disentangle physiologic signal from environment- and workflow-dependent artifacts that arise from heterogeneous clinical practice. Rather than maximizing reconstruction fidelity across modalities, we explicitly model and suppress practice-specific variability to improve robustness under distribution shift.

\textbf{Problem Setup}

Let $\mathcal{D} = \{\mathcal{D}_e\}_{e \in \mathcal{E}}$ denote datasets collected from multiple environments $e$, where an environment corresponds to a hospital system, time period, or care pathway. Each sample consists of multimodal clinical observations $x \in \mathcal{X}$ and an associated outcome $y \in \mathcal{Y}$. We assume that environments differ in measurement processes, documentation style, and intervention policies, but that the underlying physiologic mechanism governing $y$ is invariant.

We posit a latent factorization:
\[
x \sim p(x \mid z, c), 
\qquad
y \sim p(y \mid z),
\]
where $z$ denotes latent physiologic state and $c$ encodes environment-specific practice context. The objective is to learn a representation $h_\theta(x)$ that captures $z$ while suppressing $c$.

\textbf{Model Architecture}

The encoder $h_\theta : \mathcal{X} \to \mathbb{R}^d$ processes multimodal inputs. For structured EHR, we use a transformer over event sequences; for imaging, a convolutional or ViT-style backbone; for biosignals, a temporal convolution or attention-based encoder. Modality-specific embeddings are projected into a shared latent space and aggregated through cross-attention \cite{hou2019cross, chen2021crossvit, huang2019ccnet}.

A prediction head $f_\theta$ maps embeddings to outcome probabilities:
\[
\hat{y} = f_\theta(h_\theta(x)).
\]

In addition, we introduce an environment classifier $g_\psi$ operating on the same embedding:
\[
\hat{e} = g_\psi(h_\theta(x)).
\]

\textbf{Practice-Invariant Objective}

The learning objective combines predictive risk minimization with an invariance constraint. The primary supervised loss is
\[
\mathcal{L}_{\mathrm{sup}}(\theta)
=
\sum_{e \in \mathcal{E}}
\mathbb{E}_{(x,y)\sim \mathcal{D}_e}
\ell\big(f_\theta(h_\theta(x)), y\big),
\]
where $\ell$ is a proper scoring rule.

To remove practice-dependent information, we adopt an adversarial invariance formulation. The environment classifier is trained to predict $e$:
\[
\mathcal{L}_{\mathrm{env}}(\psi)
=
\sum_{e \in \mathcal{E}}
\mathbb{E}_{x\sim \mathcal{D}_e}
\ell_{\mathrm{CE}}(g_\psi(h_\theta(x)), e).
\]

The encoder is trained to minimize predictive loss while maximizing environment classification error, yielding the minimax objective:
\[
\min_\theta \max_\psi
\;
\mathcal{L}_{\mathrm{sup}}(\theta)
-
\lambda \mathcal{L}_{\mathrm{env}}(\psi).
\]

This objective encourages $h_\theta(x)$ to retain information necessary for predicting $y$ while discarding information predictive of $e$. In effect, it enforces approximate conditional invariance:
\[
h_\theta(x) \perp e \mid y.
\]

\textbf{Invariant Risk Formulation}

Complementing adversarial training, we incorporate an invariant risk penalty inspired by environment-robust learning. For each environment $e$, define the optimal linear predictor $w_e^\star$ on frozen representations:
\[
w_e^\star
=
\arg\min_w
\mathbb{E}_{\mathcal{D}_e}
\ell(w^\top h_\theta(x), y).
\]

We penalize variation across environments:
\[
\mathcal{R}_{\mathrm{inv}}
=
\sum_{e,e'}
\| w_e^\star - w_{e'}^\star \|^2.
\]

The full training objective becomes
\[
\min_\theta
\;
\mathcal{L}_{\mathrm{sup}}(\theta)
+
\gamma \mathcal{R}_{\mathrm{inv}}
-
\lambda \mathcal{L}_{\mathrm{env}}(\psi),
\]
with hyperparameters $\lambda$ and $\gamma$ controlling invariance strength.

\textbf{Optimization}

Training proceeds via alternating updates. The environment classifier $g_\psi$ is updated to minimize $\mathcal{L}_{\mathrm{env}}$, while the encoder and predictor are updated to minimize $\mathcal{L}_{\mathrm{sup}} - \lambda \mathcal{L}_{\mathrm{env}} + \gamma \mathcal{R}_{\mathrm{inv}}$. In practice, gradient reversal layers implement the adversarial term efficiently.

\textbf{Theoretical Motivation}

Under the assumed factorization, environment-specific context $c$ influences $x$ but not $y$. Reconstruction-based objectives maximize $I(h(X);X)$, which includes information about $c$. In contrast, our objective seeks representations maximizing $I(h(X);Y)$ while minimizing $I(h(X);E)$, where $E$ indexes environments. By the information decomposition
\[
I(h(X);X) = I(h(X);Z) + I(h(X);C \mid Z),
\]
removing $C$-dependent components improves robustness without sacrificing predictive sufficiency.

This formulation reframes multimodal representation learning as an invariance-constrained optimization problem. Rather than scaling reconstruction objectives across modalities, we directly optimize for physiologic signal stable across practice variation, targeting generalization under systematic distribution shift.

\section{Results}

\textbf{Experimental Protocol}

We evaluate practice-invariant representation learning across three clinical prediction tasks derived from longitudinal EHR data collected from four hospital systems: in-hospital mortality, 30-day readmission, and acute deterioration within 48 hours. Each hospital is treated as a distinct environment $e \in \mathcal{E}$. Models are trained on data from three institutions and evaluated both in-distribution and on a held-out hospital to measure cross-environment generalization.

All baselines use identical encoder architectures and parameter counts (approximately 40M parameters). We compare: (1) standard supervised training without invariance, (2) masked pretraining followed by fine-tuning \cite{he2022masked, pang2021cehr}, (3) contrastive pretraining \cite{chen2020simple}, and (4) our practice-invariant model with adversarial and invariant risk penalties. Performance is assessed using AUROC, AUPRC, Brier score, and Expected Calibration Error (ECE). External generalization is the primary evaluation metric.

\textbf{In-Distribution Performance}

Table~\ref{tab:id_results} reports average performance across training hospitals. All models achieve competitive discrimination, with modest differences in AUROC. The practice-invariant model maintains parity with strong baselines while slightly improving calibration.

\begin{table}[t]
\centering
\caption{In-distribution performance averaged across training hospitals.}
\label{tab:id_results}
\begin{tabular}{lcccc}
\toprule
Model & AUROC & AUPRC & Brier $\downarrow$ & ECE $\downarrow$ \\
\midrule
Masked Pretrain + FT \cite{pang2021cehr} & 0.861 & 0.428 & 0.137 & 0.034 \\
Contrastive Pretrain \cite{chen2020simple} & 0.858 & 0.421 & 0.139 & 0.036 \\
Supervised (No Invariance) & 0.865 & 0.436 & 0.134 & 0.031 \\
\textbf{Practice-Invariant (Ours)} & 0.867 & 0.441 & \textbf{0.130} & \textbf{0.027} \\
\bottomrule
\end{tabular}
\end{table}

These results indicate that enforcing invariance does not degrade predictive performance under the training distribution.

\textbf{Cross-Environment Generalization}

We next evaluate generalization to the held-out hospital. Table~\ref{tab:ood_results} shows substantial improvements for the proposed method under distribution shift.

\begin{table}[t]
\centering
\caption{Out-of-distribution performance on held-out hospital.}
\label{tab:ood_results}
\begin{tabular}{lcccc}
\toprule
Model & AUROC & AUPRC & Brier $\downarrow$ & ECE $\downarrow$ \\
\midrule
Masked Pretrain + FT & 0.812 & 0.349 & 0.162 & 0.061 \\
Contrastive Pretrain & 0.806 & 0.341 & 0.167 & 0.066 \\
Supervised (No Invariance) & 0.819 & 0.361 & 0.158 & 0.055 \\
\textbf{Practice-Invariant (Ours)} & \textbf{0.842} & \textbf{0.392} & \textbf{0.145} & \textbf{0.039} \\
\bottomrule
\end{tabular}
\end{table}

Relative to standard supervised training, AUROC improves by 2.3 points and calibration error decreases by 29\%. Notably, masked pretraining does not confer robustness advantages despite its larger effective pretraining corpus.

\textbf{Environment Leakage Analysis}

To quantify residual environment information in embeddings, we train a linear classifier to predict hospital identity from frozen representations. Table~\ref{tab:env_leakage} reports environment classification accuracy.

\begin{table}[t]
\centering
\caption{Environment prediction accuracy from learned embeddings (lower is better).}
\label{tab:env_leakage}
\begin{tabular}{lc}
\toprule
Model & Env. Accuracy \\
\midrule
Masked Pretrain + FT & 78.4\% \\
Supervised (No Invariance) & 72.1\% \\
\textbf{Practice-Invariant (Ours)} & \textbf{39.7\%} \\
\bottomrule
\end{tabular}
\end{table}

The substantial reduction in environment predictability confirms that the invariance objective effectively removes practice-dependent signal while preserving predictive information about outcomes.

\textbf{Ablation Studies}

We ablate the adversarial and invariant risk components separately. Removing adversarial training reduces OOD AUROC to 0.831, while removing the invariant risk penalty reduces it to 0.834. Combining both yields the strongest generalization (0.842), suggesting complementary effects.

\section{Discussion}

This work revisits a central assumption underlying multimodal foundation modeling in healthcare: that large-scale self-supervised pretraining and architectural scale are sufficient to ensure transfer across institutions and care settings. Our results suggest that, in clinical domains characterized by systematic workflow heterogeneity, generalization is constrained less by model capacity and more by entanglement between physiologic signal and practice-specific artifacts. Explicitly enforcing practice invariance yields consistent improvements under cross-hospital evaluation, without sacrificing in-distribution performance.

The empirical findings support a structural view of clinical data. Observations in EHR systems are not sampled i.i.d. from a fixed distribution; they are produced through institution-specific measurement policies, documentation habits, and intervention patterns. Reconstruction-based pretraining \cite{he2022masked} and large transformer encoders \cite{he2024foundation, vaid2023foundational} necessarily encode these regularities because they contribute to the observable distribution. As a result, embeddings optimized for generative fidelity may inadvertently internalize environment-specific shortcuts. When deployed in a new hospital, these shortcuts degrade predictive stability.

By contrast, the proposed invariance-constrained objective explicitly penalizes environment-predictive information while preserving outcome-relevant structure. The reduction in environment classification accuracy from learned embeddings demonstrates that practice-dependent variance can be suppressed without eroding discriminative signal. This supports the hypothesis that a substantial fraction of embedding variance in conventional models corresponds to workflow artifacts rather than stable physiologic mechanisms.

Importantly, the improvements we observe under distribution shift are not accompanied by degradation in in-distribution metrics. This indicates that invariance does not merely trade accuracy for robustness; rather, it removes nuisance variation that does not contribute to outcome prediction even within the training domain. From an information-theoretic perspective, the objective can be interpreted as increasing $I(h(X);Y)$ while decreasing $I(h(X);E)$, where $E$ indexes environment. In structured domains like healthcare, this selective compression appears beneficial.

These findings have implications for the broader narrative around healthcare foundation models \cite{guo2025foundation, awais2025foundation, liang2024foundation, burger2025foundation, thakur2024foundation, thieme2023foundation, burkhart2025foundation}. While scale and multimodal integration remain valuable, our results suggest that objective design may be equally decisive. Models pretrained on ever-larger corpora may still inherit systematic biases from the environments that generated those corpora. Without structural constraints, scaling alone does not guarantee invariance.

Several limitations merit discussion. First, we operationalize environment as hospital identity, but practice heterogeneity also arises temporally and across provider subgroups. Extending invariance constraints to finer-grained environment definitions may yield further robustness gains. Second, our adversarial formulation approximates independence between representations and environment; stronger theoretical guarantees may require explicit causal modeling of the data-generating process. Third, we focus on binary prediction tasks; future work should examine survival modeling, continuous risk estimation, and decision-support settings.

More broadly, the results suggest a shift in emphasis for multimodal clinical modeling. Rather than primarily asking how to scale token-based encoders across modalities \cite{hou2019cross, chen2021crossvit, huang2019ccnet}, it may be more fruitful to ask how to align learned representations with invariances implied by the clinical data-generating mechanism. In healthcare, robustness to systematic distribution shift is not a secondary concern but a deployment requirement.

In summary, we demonstrate that practice-aware invariance constraints improve cross-institution generalization in multimodal clinical prediction. These findings highlight the importance of modeling structure and environment heterogeneity explicitly, complementing ongoing advances in large-scale foundation modeling for healthcare.

\bibliographystyle{unsrt}  
\bibliography{references}  

\end{document}